\documentclass{article} % For LaTeX2e
\usepackage{iclr2023_workshop,times}

\usepackage{amsmath,amsfonts,bm}

\def\eqref#1{equation~\ref{#1}}
\def\1{\bm{1}}

\DeclareMathAlphabet{\mathsfit}{\encodingdefault}{\sfdefault}{m}{sl}
\SetMathAlphabet{\mathsfit}{bold}{\encodingdefault}{\sfdefault}{bx}{n}

\DeclareMathOperator*{\argmin}{arg\,min}

\usepackage{url}
\usepackage{graphicx}
\usepackage{wrapfig}
\usepackage{amssymb}
\usepackage{textcomp}
\usepackage{graphicx,stackengine,scalerel}
\def\tang{\ThisStyle{\abovebaseline[0pt]{\scalebox{-1}{$\SavedStyle\perp$}}}}
\usepackage{booktabs}
\usepackage{natbib}

\definecolor{generalcite}{rgb}{0.0156, 0.0745, 0.4196}
\usepackage[pdftitle={Physics-constrained neural differential equations for learning multi-ionic transport},colorlinks=true,citecolor=generalcite,urlcolor=generalcite]{hyperref}

\title{Physics-constrained neural differential  \hbox{}\hfill\linebreak equations for learning multi-ionic transport}

\author{Danyal Rehman \\
Center for Computational Science and Engineering\\
Massachusetts Institute of Technology\\
Cambridge, MA 02139, USA \\
\texttt{\href{mailto:drehman@mit.edu}{\textnormal{drehman@mit.edu}}} \\
\And
John H.\ Lienhard \\
Department of Mechanical Engineering\\
Massachusetts Institute of Technology\\
Cambridge, MA 02139, USA \\
\texttt{\href{mailto:lienhard@mit.edu}{\textnormal{lienhard@mit.edu}}} \\
}

\let\tanh\relax
\DeclareMathOperator{\tanh}{tanh} 

\iclrfinalcopy % Uncomment for camera-ready version, but NOT for submission.
\begin{document}
%\vspace*{-20px}
\maketitle
%\vspace*{-10px}
\begin{abstract}\small
Continuum models for ion transport through polyamide nanopores require solving partial differential equations (PDEs) through complex pore geometries. Resolving spatiotemporal features at this length and time-scale can make solving these equations computationally intractable. In addition, mechanistic models frequently require functional relationships between ion interaction parameters under nano-confinement, which are often too challenging to measure experimentally or know \textit{a priori}. In this work, we develop the first physics-informed deep learning model to learn ion transport behaviour across polyamide nanopores. The proposed architecture leverages neural differential equations in conjunction with classical closure models as inductive biases directly encoded into the neural framework. The neural differential equations are pre-trained on simulated data from continuum models and fine-tuned on independent experimental data to learn ion rejection behaviour. Gaussian noise augmentations from experimental uncertainty estimates are also introduced into the measured data to improve model generalization. Our approach is compared to other physics-informed deep learning models and shows strong agreement with experimental measurements across all studied datasets. 
\end{abstract}\normalsize

\section{Introduction}
Highly-selective polyamide membranes are used ubiquitously across the separations industry to recover valuable metals, such as lithium and cobalt \citep{DuChanois2023Prospects}. Owing to the rapid growth of the electric vehicle industry, the demand for these metals is expected to double by 2025 and quadruple by 2030 \citep{doi:10.1126/science.aaz6003}. To meet this increasing demand, optimizing the selectivity of polyamide membranes across diverse water sources is essential and of substantial industrial interest. Models that accurately predict ion transport and selectivity without the need for expensive experiments can play a significant role in achieving these objectives \citep{doi:10.1126/science.abm7192}. \\ \\
Continuum dynamics models are frequently used to describe the underlying laws of physical phenomena using partial differential equations (PDEs) \citep{NEURIPS2020_4b21cf96, RAISSI2019686, brunton2016discovering}. Solving these PDEs yields high solution accuracy but often at the expense of computational cost \citep{Karniadakis2021Nature}. With ion transport across highly-selective polyamide nanopores, which are often fabricated using chaotic chemical processes like interfacial polymerization, classical numerical methods encounter highly irregular and non-homogenous boundary conditions that necessitate high-resolution discretizations to resolve \citep{ritt2020ionization,jimenezsolomon2016polymer}. Additionally, at these length-scales, ion interactions become non-negligible and require functional relationships between interaction parameters and the governing PDE -- under nano-confinement, these typically invariant parameters start to diverge from bulk values due to the spatial orientation constraints \citep{10.1038/s41565-020-0713-6,doi:10.1021/acsmaterialslett.2c00932, GEISE20141}\footnote{Additional research pertaining to deep learning for PDEs and continuum ion transport models is covered in Appendix \ref{sec:relevantresearch}.}. \\ \\
To address these concerns, physics-informed neural solvers that abstract out the nature of these complex boundaries and pore geometries can prove meaningful in deriving accurate and generalizable models for ion transport under nano-confinement. In this work, we propose a physics-constrained architecture that combines neural differential equations with hard inductive biases to account for charge conservation \citep{chen2018neural}. In addition, we leverage established mechanistic solvers and independent experimental data to train the neural architecture to improve generalization across diverse concentration inputs \citep{membranes11020128,BOWEN20021121}. Gaussian noise augmentations are also introduced to the measured data to improve solver performance. This effort is the first attempt at using physics-constrained deep learning to learn complex ion transport dynamics and rejection behaviour across highly-selective polyamide membranes.  

\section{Proposed Architecture}
\paragraph{Neural Ordinary Differential Equations}The continuous dynamics of the hidden layers, $\textbf{h}$, in the neural network are parameterized using a first-order ordinary differentiatial equation (ODE):\
\begin{equation}
\frac{d{\textbf{h}(J_v)}}{dJ_v} = f_{\theta}(\textbf{h}(J_v), J_v; \theta)
\end{equation}

\noindent for flux, $J_v = \{0 \dots \mathcal{J}_v\}$, and $\textbf{h} \in \mathbb{R}^d$, where $d$ denotes the maximum number of ionic species present across all datasets. Additionally, $f_{\theta}\hspace{-4px}: [0, \mathcal{J}_v] \times \mathbb{R}^d \rightarrow \mathbb{R}^d$. To account for mixtures with different ions in the training and test data, masking is applied to $\textbf{h}(0)$. Here, outputs of the hidden layer correspond to scalar concentrations, $\textbf{h}(J_v)$. Additionally, $\theta \in \Theta$, where $\Theta$ represents some finite dimensional parameter space \citep{chen2018neural}. By learning the derivative of the hidden layer output, concentrations are uniformly Lipschitz continuous in $\textbf{h}(J_v)$
and continuous in $J_v$, enabling pre-training on classical continuum models \citep{https://doi.org/10.48550/arxiv.2202.02435}. \\ \\
In addition to masking, polynomial positional encodings are used. The embeddings are concatenated with the masked concentration vector prior to being passed into the neural network. To integrate over the neural ODE, we use the Dormand--Prince 5(4) numerical method \citep{CALVO199091} and backpropagate through the solver using the continuous adjoint method \citep{chen2018neural}. \\ \\
The network is comprised of five linear layers and ${\tanh}(\cdot)$ non-linearities applied to each output. Following the last linear layer, no point-wise activations are used. The network is trained using Adam with a batch size of 32 and an initial learning rate of $10^{-3}$ \citep{https://doi.org/10.48550/arxiv.1412.6980}. The learning rate is halved every 200 epochs for a total of 1000 epochs. For all experiments conducted, we evaluate the hidden state dynamics and their derivatives on the GPU using PyTorch, which were obtained from Python's {scipy.integrate} package \citep{SciPy2020Nature,NEURIPS2019_bdbca288}. 

\subsection{Inductive Biases}
\paragraph{Charge Conservation} Electroneutrality in ion-fluid systems is a conservation law that necessitates a solution's net charge remain neutral under equilibrium conditions \citep{doi:10.1021/acs.chemrev.1c00396}. Within the nanopores, local electroneutrality can break down \citep{PhysRevE.104.044803};\ however, in the bulk fluid, $\forall J_v$, the following constraint holds:\
\begin{equation}
\sum_{j = 1}^{d} z_j \textbf{h}_j(J_v) = 0
\end{equation}

\noindent where $z \in \mathbb{Z}^d$ is a vector of valences. To encode electroneutrality into the neural network as a hard constraint, we evaluate the orthogonal projection of the hidden layer output:\ $z^{\tang}\textbf{h}_{\perp} = z^{\tang} \textbf{h} - z^{\tang} \textbf{h}_{\parallel}$. By using $\textbf{h}_{\perp}$ instead of appending the inductive bias to the loss as a soft constraint, the model enforces inter-ionic coupling between ions, substantially improving generalization performance.

\subsection{Training and Augmentations}
\paragraph{Pre-training on Continuum Models}To pre-train the neural architecture, we use simulated data generated from the well-established Donnan-Steric Pore Model with Dielectric Exclusion (DSPM-DE). We use an iterative, under-relaxed numerical scheme to solve the PDE \citep{GERALDES2008172} (model and implementation details are provided in Appendix \ref{sec:DSPMDE}):\ 
\begin{equation}
J_i = -D_iK_{i, d}\partial_x C_i + K_{i, c}C_iJ_v - \frac{K_{i, d}D_i C_i z_i {F}}{RT}\partial_x \psi, \hspace*{20px} x \in [0, \Delta x_e]
\end{equation}

\noindent In DSPM-DE, four latent variables are most often used to parameterize the nanoporous membrane:\ $\mathcal{Z} = \{r_p, \Delta x_e, \zeta_p, \chi_d\}$. We apply our previously-developed approach using global optimization with simulated annealing and the Nelder-Mead local search to regress average values of the four parameters across the training data \citep{REHMAN2022100034}. Details are provided in Appendix \ref{sec:DSPMDE}.\\ \\%\vspace{-10px}
During pre-training, the $d$-dimensional concentration vector was sampled using low-discrepancy Sobol sequences and projected to log-space to improve model predictions at lower concentrations \citep{joe2008constructing,WANG202116906}. The MSE loss used for pre-training is:\
\begin{equation}
\mathcal{L}_{\mathrm{cm}}(\textbf{h}, \textbf{h}^{\mathrm{cm}}) = \frac{1}{kd}\sum_{i = 1}^k\sum_{j = 1}^d \left[\textbf{h}_{j}(J_{v, i}) - {\textbf{h}^{\mathrm{cm}}_{j}}(J_{v, i})\right]^2
\end{equation}

\paragraph{Fine-tuning on Measured Data}
Following pre-training, the network was fine-tuned using experimental data comprising 850 ion concentration measurements. To improve generalization, we use measured uncertainties across the training data to fit Gaussian statistics to individual data points. To evaluate the loss function, values of output concentration are sampled from this distribution\footnote{Hinge loss terms based on the Hofmeister series were originally included in the loss function but provided mixed results \citep{SOMRANI2013184}. Given that ion rejection has been seen to diverge from the Hofmeister series under certain conditions, it was removed entirely from the loss function \citep{LUO2011145}.}:\
\begin{equation}
\mathcal{L}_{\mathrm{exp}}(\textbf{h}, \textbf{h}^{\mathrm{exp}}) = \frac{1}{nd}\sum_{i = 1}^n\sum_{j = 1}^d \left[\textbf{h}_{j}(J_{v, i}) - {\textbf{h}^{\mathrm{exp}}_{j}}(J_{v, i})\right]^2, \hspace{10px}{\textbf{h}_{j}^{\mathrm{exp}}}(J_{v, i}) \sim \mathcal{N}(\mu_{ij}, \sigma_{ij}^2) \ \forall i,j
\end{equation}

\noindent All datasets analyzed the same polyamide membrane (DuPont's FilmTec{\texttrademark} NF270) for ion separation \citep{ALZOUBI200742,MICARI2020118117,Zoubi2009Nanofiltration,doi:10.1021/acs.est.7b06400}. To weight the measured data more heavily than the simulated data during training, $n \ll k$. 

\section{Results and Discussion}
\begin{figure}[h]
\begin{center}
%\framebox[4.0in]{$\;$}
%\fbox{\rule[-.5cm]{0cm}{4cm} \rule[-.5cm]{4cm}{0cm}}
\includegraphics[width=\textwidth]{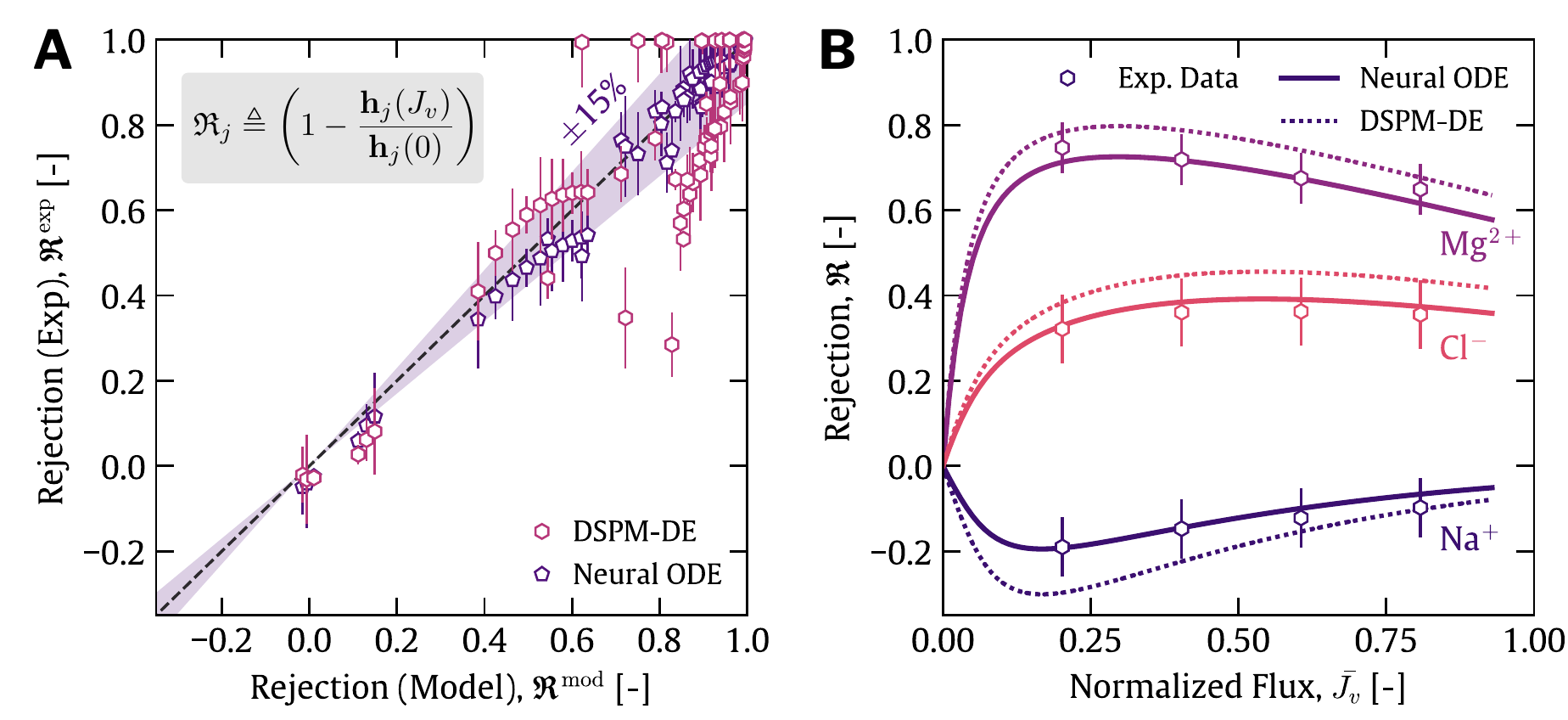}
\end{center}
\caption{\textbf{Left}:\ Parity plot illustrating superior predictive performance of the physics-constrained neural ODE over the classical mechanistic solver across all test data. \textbf{Right}:\ For a given test composition, ion rejection is predicted as a function of normalized flux by the neural solver and DSPM-DE.}
\label{fig:ODENetvDSPM}
\end{figure}

\paragraph{Downstream Task Prediction}To evaluate the performance of our framework, we compare ion rejection predictions with those from the continuum model and measurement data in the test set (Fig.\ \ref{fig:ODENetvDSPM}A). Rejection values closer to 1 signify perfect selectivity between the ion and the polyamide membrane. Ion rejection can also undertake values below 0, which signify counter-ion entrainment through the pore to satisfy the electroneutrality condition in the product flow \citep{GILRON2001223}. Negative rejection is frequently observed in ion transport across polyamide nanopores \citep{doi:10.1021/acs.chemrev.1c00396}, and here, we illustrate the neural solver's ability to capture this phenomena and generalize it more accurately than the baseline DSPM-DE method. \\ \\
In Fig.\ \ref{fig:ODENetvDSPM}B, a sample initial condition from the test set is propagated through the neural solver and DSPM-DE. Since the baseline method requires knowledge of the hydrated ionic radii and membrane charge (a parameter with high sensitivity to local concentration and composition) $-$ both of which can possess large uncertainties $-$ error in these parameters propagates into ion rejection \citep{REHMAN2022100034}. Since the neural ODE eliminates the need for these parameters by treating the polyamide like a black-box, predictive performance is substantially improved. \\
\vspace{-10px}

\paragraph{Other Architectures}We compare our approach to other deep learning methods and mechanistic models in Table \ref{tab:NNcomparison}. We test physics-informed neural networks (PINNs)\footnote{To evaluate the PDE loss in the PINN case, DSPM-DE was used to solve the Nernst-Planck equations.} \citep{RAISSI2019686}, ResNets \citep{https://doi.org/10.48550/arxiv.1512.03385}, and U-Nets \citep{https://doi.org/10.48550/arxiv.1505.04597}. For the mechanistic models, we use DSPM-DE as the baseline method \citep{BOWEN20021121}. We also evaluate the solution-friction (SF) method on the test data using ion-specific friction factors regressed from the training set \citep{doi:10.1021/acs.est.1c05649}. Since inter-species diffusion coefficients for all ions in the training and test data were unavailable in the literature, the Maxwell--Stefan framework could not be used for comparison purposes. The projection operator, noise augmentations, and training procedure were all held constant for the comparison, as were the approximate number of parameters used in the deep learning models. Hyperparameter tuning was also performed on all the learning architectures to ensure a fair comparison between rejection predictions.\\
\begin{table}[h]\small
\caption{Downstream task test error using alternate deep learning methods and continuum models.}
\label{tab:NNcomparison}
\begin{center}
\begin{tabular}{ccc}\toprule
\multicolumn{1}{c}{}  &\multicolumn{1}{c}{\# Params}&\multicolumn{1}{c}{Test Error}
\vspace{2px} \\ \midrule 
\textbf{Deep Learning Methods} &  &   \\
Physics-constrained Neural ODE (Our Work)& 1.05 M & 7.1\%  \\
PINN \citep{RAISSI2019686} & 1.12 M & 9.7\%  \\
ResNet \citep{https://doi.org/10.48550/arxiv.1512.03385} & 1.23 M & 9.2\% \\
U-Net \citep{https://doi.org/10.48550/arxiv.1505.04597} & 1.01 M & 10.2\%  \\ \midrule
\textbf{Continuum Models} &  &   \\
DSPM-DE (baseline) \citep{BOWEN20021121} & 4 & 17.8\%  \\
SF \citep{doi:10.1021/acs.est.1c05649} & $d$ + 4 & 24.8\%  \\
 \bottomrule
\end{tabular}
\end{center}
\end{table}\vspace{-10px}
%\begin{table}[h]\small
%\caption{Downstream task test error using alternate deep learning methods and continuum models.}
%\label{tab:NNcomparison}
%\begin{center}
%\begin{tabular}{ccc}\toprule
%\multicolumn{1}{c}{}  &\multicolumn{1}{c}{\# Params}&\multicolumn{1}{c}{Test Error}
%\vspace{2px} \\ \midrule 
%\textbf{Deep Learning Methods} &  &   \\
%Neural ODE (Our Work)& 1.05 M & 7.1\%  \\
%FNO \citep{https://doi.org/10.48550/arxiv.2010.08895} & 1.17 M & 9.3\%  \\
%PINN \citep{RAISSI2019686} & 1.12 M & 10.4\%  \\
%ResNet \citep{https://doi.org/10.48550/arxiv.1512.03385} & 1.23 M & 10.7\% \\ \midrule
%\textbf{Continuum Models} &  &   \\
%DSPM-DE (baseline) \citep{BOWEN20021121} & 4 & 17.8\%  \\
%SF \citep{doi:10.1021/acs.est.1c05649} & $d$ + 4 & 24.8\%  \\
% \bottomrule
%\end{tabular}
%\end{center}
%\end{table}\vspace{-10px}

\noindent Across the test data, we find that the physics-constrained neural ODE outperforms the PINN, ResNet, and U-Net. This is largely attributed to the smooth concentrations predicted by the neural ODE, which are advantageous in bounding test error when unseen concentrations and fluxes are observed \citep{https://doi.org/10.1007/s10915-022-01939-z}. Despite this, the other models still exhibit lower test error than the baseline DSPM-DE method, reiterating the importance of relaxing the simplifications encoded into classical mechanistic models. Similarly, in agreement with expectation, the SF method produces the largest error on the test set as a result of the overparameterized friction factors (see Appendix \ref{sec:relevantresearch} for details). Despite the fewer latent variables in DSPM-DE, it still exhibits higher prediction accuracy than the SF model. Overall, the neural ODE provides the best performance in predicting ion transport across the polyamide nanopores across diverse input concentrations.

\section{Conclusions and Future Work}
In this work, we develop the first physics-constrained neural ODE solver for ion transport across polyamide nanopores. Our solver encodes electroneutrality into the architecture and trains on a mixture of simulated data and experimental measurements augmented with Gaussian noise to achieve an average test error of 7.1\%, compared to 17.8\% from classical PDE solvers. The model also outperforms other deep learning models when rejection prediction is used as the downstream task. Next steps include generating new ion rejection profiles using complex mixtures to identify high-value separations using polyamide membranes. Additionally, we hope to quantify the model's predictive accuracy in the high-salinity regime for applications to metal recovery from hypersaline brines.

\newpage\clearpage
\subsubsection*{Acknowledgments}
The authors thank the Centers for Mechanical Engineering Research and Education at MIT and SUSTech (MechERE Centers at MIT and SUSTech) for partially funding this research. D.R.\ acknowledges financial support provided by a fellowship from the Abdul Latif Jameel World Water and Food Systems (J-WAFS) Lab and fellowship support from the Martin Family Society of Fellows.

\small
\bibliography{iclr2023_workshop}
\bibliographystyle{iclr2023_workshop}

\normalsize
\newpage\clearpage
\appendix
\section{Appendix}
\subsection{Relevant Work}
\label{sec:relevantresearch}
\paragraph{Machine Learning for PDEs}A substantial amount of literature is dedicated to using neural networks to learn PDEs \citep{RAISSI2019686,RAISSI2018125,long2018pde,bar2019learning}. These efforts include applications to fluid simulation \citep{https://doi.org/10.1111/cgf.13620}, flame propagation \citep{doi:10.1073/pnas.2101784118,https://doi.org/10.48550/arxiv.2202.07643}, and shallow wave dispersion \citep{https://doi.org/10.48550/arxiv.2202.07643}. Some approaches investigate finite-dimensional solution operators to learn the governing PDE, but intrinsically depend on domain geometry and spatiotemporal discretization \citep{ZHU2018415}. Other work has considered learning infinite-dimensional neural operators for mesh-independent applications \citep{https://doi.org/10.48550/arxiv.2010.08895, Karniadakis2021Learning}. Specific to ion transport, prior research has focused on the development of deep learning models for membrane fouling prediction \citep{DEJAEGHER2021118028} and multi-scale transport \citep{RALL2020118208};\ however, studies investigating multi-ion transport across polyamide nanopores remain elusive, and the primary focus of this work.

\paragraph{Continuum Ion Transport Models}The Donnan-Steric Pore Model with Dielectric Exclusion (DSPM-DE) is one of the most frequently used models for quantifying ion transport across polyamide nanopores \citep{BOWEN20021121}. DSPM-DE solves the extended Nernst-Planck PDEs in conjunction with electroneutrality constraints that serve as a closure model for the electric potential. Other variants close the PDE using solutions to the Poisson-Boltzmann equations \citep{doi:10.1137/140968082}. In DSPM-DE, the boundary conditions are highly simplified, assuming cylindrical nanopores and equilibrium partitioning relationships to quantify selectivity mechanisms \citep{REHMAN2022100034}. The solution-friction (SF) model similarly solves the extended Nernst-Planck PDEs;\ however, it makes no assumptions about pore geometry and regresses friction factors from data to quantify restricted transport within the nanopores \citep{doi:10.1021/acs.est.1c05649}. Although the model eliminates assumptions around pore structure, regressing friction factors often overparameterizes the model making generalization challenging. Other models like the Maxwell-Stefan framework attempt to capture inter-species coupling through experimentally-measured binary diffusion coefficients \citep{KRISHNA1997861}. Although measurable in bulk solutions, determining these coefficients inside the polymer matrix is often not feasible, meaning that the model often resorts to simplifications similar to those used in DSPM-DE and SF models \citep{REHMAN2021117171}. 

\subsection{Mechanistic Model and Parameter Estimation}
\label{sec:DSPMDE}
DSPM-DE solves the extended Nernst-Planck PDE inside the active layer of the polyamide:\
\begin{equation}
J_i = -D_iK_{i, d}\partial_x C_i + K_{i, c}C_iJ_v - \frac{K_{i, d}D_i C_i z_i {F}}{RT}\partial_x \psi, \hspace*{20px} x \in [0, \Delta x_e]
\end{equation}

\noindent Here, $C_i$ is the solute concentration, $J_i$ is the solute flux, and $\psi$ is the electric potential. $F$, $R$, and $T$ are Faraday's constant, the universal gas constant, and absolute temperature, respectively. $K_{i, c}$ and $K_{i, d}$ are convective and diffusive hindrance coefficients used to capture reductions in the bulk diffusion coefficient, $D_i$, under nano-confinement. Polynomial expressions for  $K_{i, c}$ and $K_{i, d}$ can be derived from perturbation theory solutions to the Navier-Stokes equations assuming ions behave like hard spheres in cylindrical pores, where $K_{i,c} \in [0, 1]$ and $K_{i, d} \in [0, 1]$ \citep{DEEN19871409,MAVROVOUNIOTIS1988269}. \\ \\
By removing hindered transport assumptions (i.e.\ using bulk diffusion coefficients and setting $K_{i, c} = 1$ and $K_{i, d} = 1$), the extended Nernst-Planck PDE can also be used to solve for ion transport in the boundary layer. Mass transfer correlations are also needed to calculate the Sherwood number as a function of system geometry and flow characteristics \citep{LABBAN201718}. The PDEs are solved simultaneously, with electroneutrality conditions applied to each spatial domain:\
\begin{equation}
\sum_{i = 1}^d z_i C_i = 0 \hspace{20px} \sum_{i = 1}^d z_i C_i + \chi_d(C_i) = 0
\end{equation}

\begin{figure}[h]
\begin{center}
\includegraphics[width=\textwidth]{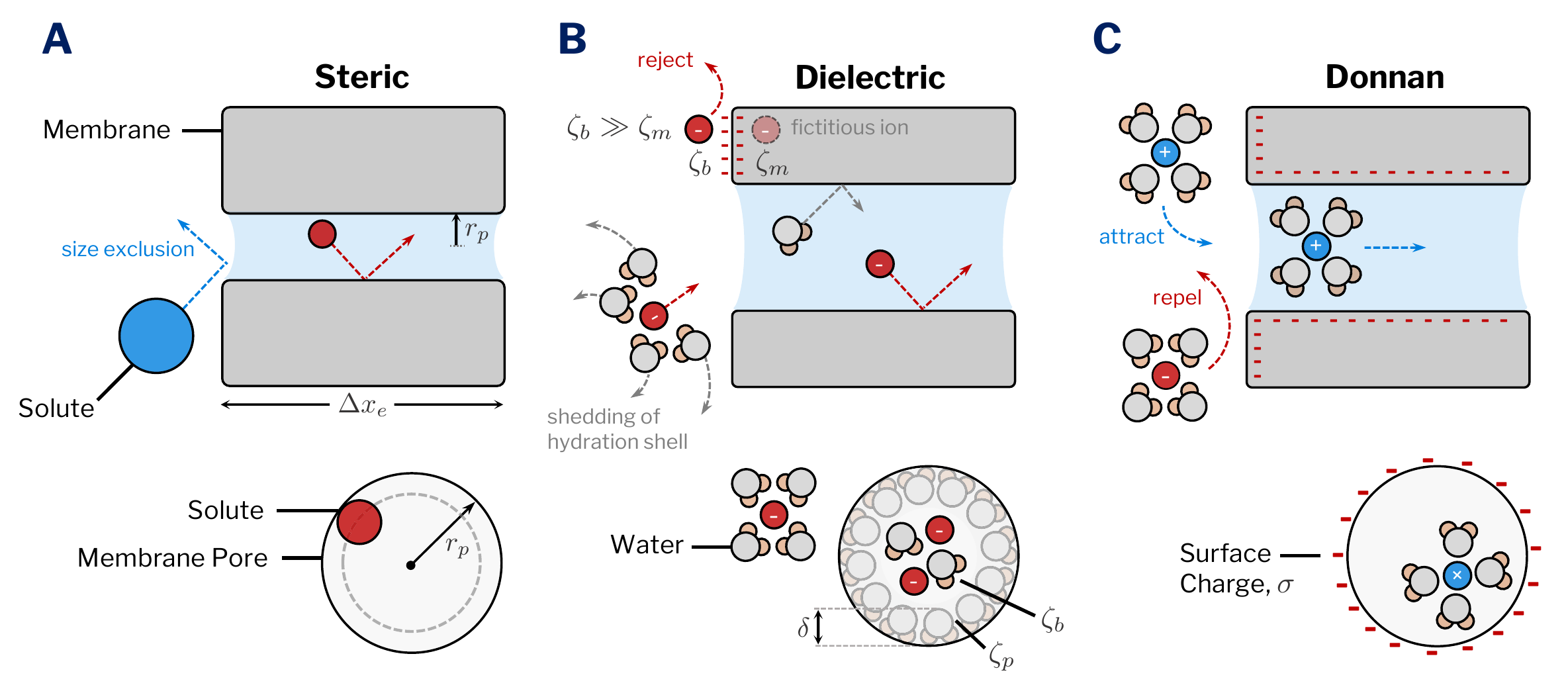}
\end{center}
\caption{Physical representation of individual ion selectivity mechanisms used to quantify boundary conditions in the DSPM-DE model. \textbf{(A)} Steric rejection separates ions based on their size relative to the pore radius, $r_p$. \textbf{(B)} Dielectric exclusion captures the partial or complete shedding of ion hydration shells prior to entering the membrane pores with fictitious image forces created at the membrane interface. Image charges are repulsive for both cations and anions, given that the dielectric constant of the solvent, $\zeta_b$, is larger than the dielectric constant of the membrane matrix, $\zeta_m$. The cross section shows a thin film of water molecules with a constrained orientation aggregated near the pore walls, where the dielectric constant of the solvent is reduced to $\zeta_p$. The thickness of the layer of water molecules is denoted by $\delta$. $\zeta_b$ and $\zeta_m$ were set to 78.54 and 4.5, respectively, in the reported work. \textbf{(C)} The Donnan exclusion mechanism fractionates ions based on charge, e.g.\ negatively-charged ions are repelled by a negatively-charged membrane, whereas positively-charged ions are attracted into the pores. }
\label{fig:SelectivyMechanisms}
\end{figure}

\noindent The first equation prescribes electroneutrality in the boundary layer film and product flow, while the second equation enforces electroneutrality inside the polyamide. $\chi_d(C)$ is the volumetric charge density of the polyamide membrane and a function of local composition \citep{doi:10.1073/pnas.2008421117}. Adsorption isotherms have been frequently used to quantify these functional dependencies. \\ \\
Lastly, boundary conditions are prescribed at the solution-membrane interface. Three equilibrium partition coefficients, $\phi_{\mathrm{S}}$, $\phi_{\mathrm{Di}}$, and $\phi_{\mathrm{Do}}$, are used to evaluate ion selectivity as a function of steric, dielectric, and Donnan exclusion terms (a schematic diagram is presented in Fig.\ \ref{fig:SelectivyMechanisms}):\
\begin{equation}
\frac{\gamma_i(0^-) C_{i}(0^-)}{\gamma_i(0^+) C_{i}(0^+)} = \phi_{i, \mathrm{S}}\phi_{i, \mathrm{Di}}\phi_{i, \mathrm{Do}}
\label{eq:BC}
\end{equation}

Here, $\gamma$ is the ion activity coefficient, which is used to capture the departure from ideality of ions in solution. In this work, the Pitzer-Kim model is used to evaluate these coefficients \citep{pitzer1973}. $0^-$ and $0^+$ denote the solution-side and membrane-side at the interface, respectively. Calculating each partition coefficient's contributions is summarized in previous work \citep{REHMAN2022100034}. 

\noindent To solve the above system of equations, two under-relaxation update schemes are used \citep{GERALDES2008172}. The first applies to the electric potential:\
\begin{equation}
\psi^{(n + 1)} = \psi^{(n - 1)} + \eta_\psi \left[\psi^{(n)} - \psi^{(n - 1)}\right]
\end{equation}

\noindent where the superscript denotes the iteration step. The under-relaxation factor, $\eta_{\psi} \in [0, 1]$. Since the governing PDE can be very stiff (due to order of magnitude differences in input concentrations), convergence of the DSPM-DE method is highly sensitive to the under-relaxation parameter. A relatively low value of $\eta_\psi = 0.10$ was used across all simulations to guarantee convergence. \\ \\
A second under-relaxation parameter is also required to ensure that ion concentrations converge. The update step is:\ 
\begin{equation}
C_{i}^{(n + 1)} = C_i^{(n - 1)}\left[1 + \eta_C\min\left(1, \bigg\lvert\frac{C_i^{(n - 1)}}{C_i^{(n)} - C_i^{(n - 1)}}\bigg\rvert\right) \right]\left(\frac{C_i^{(n)} - C_i^{(n - 1)}}{C_i^{(n - 1)}}\right)
\end{equation}

\noindent where $\eta_C \in [0, 1]$. Similarly, a low value of $\eta_C = 0.175$ was set across simulations to ensure convergence. This update formulation also guarantees that $C_i^{(n + 1)}$ remains non-negative at each iteration. \\ \\
This procedure, in conjunction with the four latent membrane variables:\ $\mathcal{Z} = \{r_p, \Delta x_e, \zeta_p, \chi_d\}$, can be used to solve for output concentrations in the product flow. To quantify the latent variables, $\mathcal{Z}$, we minimize the following objective function using a hybrid global-local optimization method:\ simulated annealing combined with the Nelder-Mead local search \citep{REHMAN2022100034}:\
\begin{equation}
\argmin_{\mathcal{Z}} \ \sum_{i = 1}^{N_s}\sum_{j = 1}^{N_w}\frac{\left[\mathfrak{R}_{i,j}^{\mathrm{mod}}(\mathcal{Z}) - \mathfrak{R}_{i,j}^{\mathrm{exp}}\right]^2}{{\sigma}_{i,j}^2}
\end{equation}
 
\noindent where $N_s$ and $N_w$ are the total number of ions and flux measurements taken, respectively. Here, $\sigma_{i, j}^2$ is the variance estimate across experimental trials. Using this approach, in Table \ref{tab:DSPMDE_Parameters}, we summarize the converged latent parameters using all data points in the training set. These values were used to generate simulated data needed for pre-training the physics-constrained neural ODE.
\begin{table}[!h]\small
\caption{DSPM-DE parameters used for pre-training the physics-constrained neural solver.}
\label{tab:classicalparams}
\begin{center}
\begin{tabular}{cccc}\toprule
\multicolumn{1}{c}{$r_p$ [nm]}  &\multicolumn{1}{c}{$\Delta x_e$ [\textmu m]}&\multicolumn{1}{c}{$\zeta_p$ [-]}&\multicolumn{1}{c}{$\chi_d$ [mol/m$^3$]}
\vspace{2px} \\ \midrule 
 0.51 & 1.27 & 43.56 & $-$51.23 \\
 \bottomrule
\end{tabular}
\label{tab:DSPMDE_Parameters}
\end{center}
\end{table}

\end{document}